\documentclass{article}

\usepackage{tcolorbox}
\usepackage{inconsolata} 
\usepackage{multirow}
\usepackage{makecell}

\usepackage[preprint]{neurips_2025}
\usepackage[utf8]{inputenc} 
\usepackage[T1]{fontenc}    
\usepackage{hyperref}       
\usepackage{url}             
\usepackage{booktabs}       
\usepackage{amsfonts}       
\usepackage{nicefrac}       
\usepackage{microtype}      
\usepackage{xcolor}         
\definecolor{pink}{RGB}{180,50,120} 

\newcommand{\pinkbold}[1]
{\texttt{\textbf{\textcolor{pink}{#1}}}}
\usepackage[none]{hyphenat}
\title{Identifying Financial Risk Information Using RAG with a Contrastive Insight}
\workshoptitle{Generative AI in Finance}

\author{
  Ali Elahi\thanks{The research was done during Summer 2025 at Surlamer Investments.} \\
  Department of Computer Science\\
  University of Illinois Chicago; Surlamer Investments\\
  Chicago, IL \\
  \texttt{aelahi6@UIC.edu} \\
}

\begin{document}

\maketitle

\begin{abstract}


In specialized domains, humans often compare new problems against similar examples, highlight nuances, and draw conclusions instead of analyzing information in isolation. When applying reasoning in specialized contexts with LLMs on top of a RAG, the pipeline can capture contextually relevant information, but it is not designed to retrieve comparable cases or related problems. 

While RAG is effective at extracting factual information, its outputs in specialized reasoning tasks often remain generic, reflecting broad facts rather than context-specific insights. In finance, it results in generic risks that are true for the majority of companies. To address this limitation, we propose a peer-aware comparative inference layer on top of RAG.

Our contrastive approach outperforms baseline RAG in text generation metrics such as ROUGE and BERTScore in comparison with human-generated equity research and risk.
\end{abstract}

\section{Introduction}
Retrieval-based systems access external documents to provide relevant information for downstream tasks \cite{lewis2021retrievalaugmentedgenerationknowledgeintensivenlp}. Retrieval-Augmented Generation (RAG) combines a similarity-based retrieval with a language model inference layer to refine and synthesize retrieved content, providing a robust framework for summarization, question answering, and information extraction. Limitation arises from RAG’s performance in retrieving critical information in specialized domains such as finance, particularly when extracting company risk factors from lengthy filings. A contextually rich passage detected by cosine similarity can still be irrelevant or uninformative, while a less rich section can convey more important details. In other words, the salience of information does not necessarily align with the semantic similarity index. We propose a contrastive approach: retrieving a broader set of relevant information for both the target firm and comparable peers, then prompting LLMs to generate a contrastive analysis that highlights nuances and distinctions.

\subsection{Financial Risk}

In financial equity research, risk identification refers to finding the key factors that could significantly affect a company’s performance, whether by helping it grow or causing it to lose value. These factors shape how analysts, investors, and decision-makers judge the company’s prospects and determine the value of its securities. A primary source for this information is the periodic reports public companies file with the U.S. Securities and Exchange Commission (SEC), including the annual document 10-K, the quarterly document 10-Q, and earnings call transcripts, records of quarterly calls where executives discuss results, highlight challenges, and answer questions from analysts, often revealing insights not found in the written filings. These documents contain sections dedicated to describing the company’s operations and financial condition.

For example, many filings include standard sections on topics like cybersecurity, regulatory compliance, or supply chain risk. These appear across most companies’ reports, regardless of their actual relevance. A retrieval system or RAG model will reliably extract these sections simply because they are explicitly labeled, even if they are low-priority or non-differentiating. The challenge is to extract the relevant information and assess its relative importance in the context of the specific company and its peers in the same sub-sector.

\section{Methodology}

\begin{figure}[h] 
    \centering
    \includegraphics[width=0.99\textwidth]{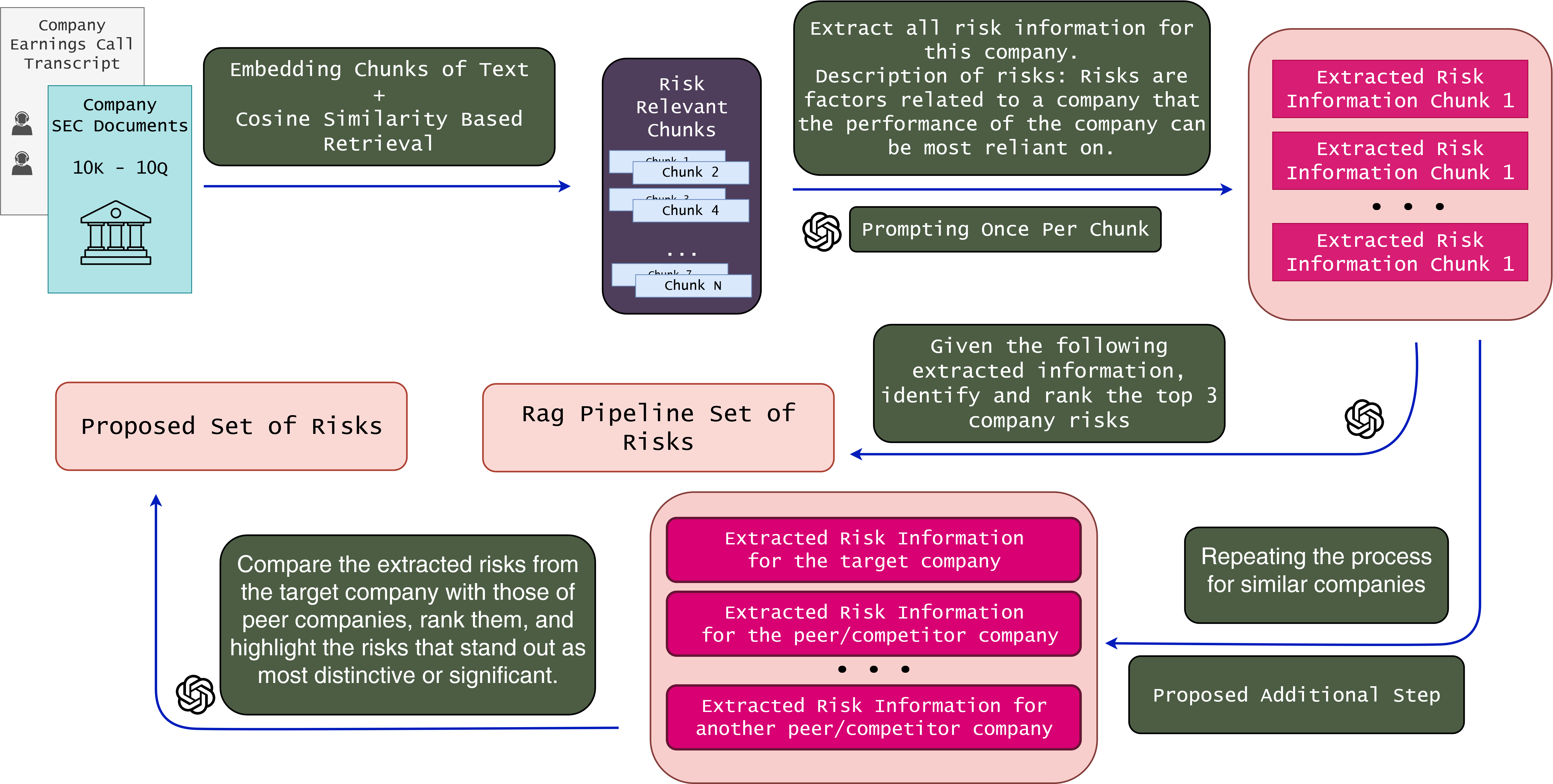} 
    \caption{Baseline approach for risk identification based on RAG and summarization mythologies generates the ``Final Set of Risks''; our proposed additional step will flag the most important items based on using a comparative approach and generate the ``Proposed Set of Risks''.}
    \label{fig:pipeline}
\end{figure}

We propose an additional stage on top of RAG to identify a company’s most significant and distinctive risks. First, the system extracts a broad set of potential risks from filings and earnings call transcripts. It then compares these with risk factors from peer companies in the same industry. By contrasting common versus unique or unusually emphasized risks, the system highlights those most critical in context.

Figure \ref{fig:pipeline} summarizes the proposed risk identification pipeline. In the first stage, text chunks most relevant to the risk query (Appendix \ref{prompt:riskquerry} – Risk Query) are extracted using cosine similarity of text embeddings. Each chunk is then processed by an LLM with a prompt (Appendix \ref{prompt:riskquerry} – Risk Information Extraction Prompt) to retrieve relevant risk information. A subsequent prompt (Appendix \ref{prompt:riskquerry} – Risk Aggregation Prompt) aggregates information across chunks, removes duplicates, and categorizes content into distinct risk topics. Finally, a contrastive prompt (Appendix \ref{prompt:riskquerry} – Contrastive Risk Identification Prompt) compares the aggregated risk information for the target company and its peers to identify risks that are most specific or significant for the target company.

In the baseline approach, after the aggregation method, we prompt the LLM to generate the most important risks for the company.

\subsection{Model sizes and Temperatures}
We used a variety of OpenAI LLMs for our experiments. The described pipeline consists of three stages: information extraction, risk aggregation, and risk identification and ranking, the latter being the stage that differs between the baseline and contrastive approaches. The first two stages use GPT-4.1-mini, the aggregation stage uses GPT-4.1, and for the final stage, we compare three models: O3 (a recent OpenAI reasoning model with reasoning level set to medium), 4o (an older OpenAI model released in late 2024), and 4.1 (OpenAI’s most capable non-reasoning model).

\subsection{Peers} Peers are companies in the same industry with similar business characteristics, such as scale (e.g., market capitalization), sector, products, or services, and whose stock returns tend to be correlated. In this work, we constructed our own peer sets for each company, though they can also be defined using methods such as GICS \cite{msci_gics} sectors or Bloomberg's peers tool.

\section{Experiments and Results}

In our analysis, we generated risks for S\&P 500 companies using the latest 10-Ks, 2025 10-Qs, and earnings call transcripts, which are all publicly available data. Relevant text chunks from these documents were extracted as input for the LLMs, and the output is a ranked list of risks.

\subsection{Evaluation Methods}
For the financial risk identification task, we evaluate performance using standard NLP summarization and text generation metrics, including ROUGE-1, ROUGE-2, ROUGE-L \cite{lin-2004-rouge}, and BERTScore \cite{zhang2020bertscoreevaluatingtextgeneration}. Instead of using the input text as the reference, we compare the outputs against the risks and insights documented in human-written equity research reports. Using human-written risks, we can compare how accurate our risks are compared to actual risks that financial analysts extracted for a company. Due to confidentiality, the research reports used in this study will not be made publicly available upon publication.

\begin{table*}[t]
\centering
\resizebox{0.88\columnwidth}!{
\begin{tabular}{ll|cc|cc|cc|cc}
\toprule
 & \textbf{} & \multicolumn{2}{|c|}{\textbf{BERTScore}} & \multicolumn{2}{c|}{\textbf{ROUGE-1}} & \multicolumn{2}{c|}{\textbf{ROUGE-2}} & \multicolumn{2}{c}{\textbf{ROUGE-L}}\\
\midrule
\textbf{Models} & \textbf{Algorithms} & \textbf{Recall} & \textbf{F1} & \textbf{Recall} & \textbf{F1} & \textbf{Recall} & \textbf{F1} & \textbf{Recall} & \textbf{F1} \\
\midrule
\multirow{2}{*}{GPT 4o}&Baseline&0.1398&0.1417& 0.1872&0.1889&0.0225&0.0239&0.0713&0.0702\\
&Contrastive&\textbf{0.1401}&\textbf{0.1436}&\textbf{0.1904}&\textbf{0.1923}&\textbf{0.0231}&\textbf{0.0246}&\textbf{0.0719}&\textbf{0.0709}\\
\midrule
\multirow{2}{*}{GPT 4.1}&Baseline&0.1504&0.1490&0.1941&0.1907&0.0228&0.0237&0.0718&0.0689\\
&Contrastive&\textbf{0.1561}&\textbf{0.1548}&\textbf{0.2177}&\textbf{0.2025}&\textbf{0.0264}&\textbf{0.0256}&\textbf{0.0783}&\textbf{0.0705}\\
\midrule
\multirow{2}{*}{GPT O3}&Baseline&0.1673&0.1570&0.1913&0.1913&0.0191&0.0201&0.0654&0.0644\\
&Contrastive&\textbf{0.1692}&\textbf{0.1605}&\textbf{0.2116}&\textbf{0.2011}&\textbf{0.0210}&\textbf{0.0210}&\textbf{0.0702}&\textbf{0.0657}\\
\bottomrule
\end{tabular}
}
\label{tab:results}
\caption{BERTScore Score Recall and F1 Scores. Recall has a higher importance since we do not want to miss any true risks.}
\end{table*}

\subsection{Results}
Table \ref{tab:results} reports the evaluation metrics for the baseline and contrastive approaches across the three OpenAI models used in the final inference layer. Across all metrics, the contrastive methodology consistently outperforms the baseline, demonstrating its ability to retrieve risks more aligned with human-written reports and investment theses. Additionally, comparing models, the O3 reasoning model achieves a higher BERTScore and ROUGE than both GPT-4o and GPT-4.1.

To better illustrate the results, we analyzed five groups of companies, each belonging to a specific sector or subsector, and examined the risk topics extracted by our pipeline. The results are presented in Table \ref{tab:example}. As shown, the identified risks align closely with the business characteristics of each sector.


\begin{table*}[t]
\centering
\resizebox{0.98\columnwidth}!{
\begin{tabular}{p{2.2cm}|p{12cm}}
\toprule
Sector/Industry & Identified Risks \\
\midrule
Oilfield Services & Commodity Price \& Customer Spending
Cyclicality; Energy Transition \& New
Technology Execution; Contractual/Operational
Execution Risk; Tariff, Trade \& Regional Market Risk Sensitivity\\
\midrule
Railing & Tariff, Trade \& Regional Market Risk Sensitivity; Product \& Innovation Execution, Raw Material Sourcing \& Seasonality; Product Quality \&
Compliance; Competitive Pricing Pressure \\
\midrule
Tech & Competitive Pricing Pressure; Technology
Platform Execution Risk; Intellectual Property Risk; FX Sensitivity \& Macro Dependence \\
\midrule
Healthcare & Patent/Exclusivity \& Product
Concentration; Regulatory, Pricing \& Market Access; Pipeline, R\&D, M\&A, and Integration Risk \\
\midrule
Financial Services & FX Sensitivity \& Macro
Dependence; Actuarial/ Underwriting/Reserve
Risk; Transition Finance \& ESG Exposure; Investment Portfolio \& Market Volatility \\
\bottomrule
\end{tabular}
}
\caption{Risk titles identified for each industry/sector.}
\label{tab:example}
\end{table*}

\section{Literature Review}
Traditionally, investment banks spent significant time extracting details from filings, earnings call transcripts, and corporate presentations before synthesizing them into reports. NLP methods dramatically accelerate this process, and various methodologies are proposed to identify risk and summarize financial information using NLP\cite{pei2024modeling, zhou2024finrobotaiagentequity, mahfouz2021framework}. Beyond supporting investment decisions, the extracted and summarized information can also serve as input to downstream LLM pipelines, such as those designed to assess stock valuations given risk and other factors \cite{zhao2025alphaagentslargelanguagemodel, 10825449}.

Recent work has combined RAG and reasoning components for general or specialized tasks. C-RAG \cite{ranaldi2024eliciting} introduces a contrastive framework in which retrieved documents are used to generate explanatory arguments, and teacher-model explanations serve as demonstrations for student models, blending retrieval with contrastive few-shot reasoning in general QA tasks. RAFT proposes a training scenario that strengthens in-domain RAG by teaching models to identify and cite relevant evidence while ignoring ``distractor'' documents, coupling retrieval with chain-of-thought reasoning to improve factual grounding in domains such as biomedicine \cite{zhang2024raft}. RankRAG \cite{yu2024rankrag} integrates retrieval ranking and answer generation within a single instruction-tuned LLM, showing that incorporating ranking signals improves both context selection and downstream reasoning, often outperforming expert rankers and strong generation baselines across multiple benchmarks. Finally, RAG+ explicitly incorporates application-aware reasoning by retrieving not only factual knowledge but also aligned application examples, enabling structured, goal-directed inference; this design yields consistent improvements across domains such as mathematics, law, and medicine\cite{wang2025rag}.

\section{Limitations}
One limitation of our pipeline is its lack of temporal awareness. The methodology does not account for the timeline of ongoing company events, meaning emerging developments are not captured, even though such events may materially affect risk exposure. As a result, the system may overlook dynamic changes that are crucial for timely and accurate risk assessment.

Another limitation lies in evaluation. Current assessments rely primarily on NLP-based text generation metrics, which measure surface-level similarity but fail to capture whether the extracted risks are truly aligned with a company’s actual exposures. A stronger evaluation framework would incorporate expert human judgment as well as reasoning-based methods that go beyond lexical overlap to assess factual accuracy and contextual relevance.

\section*{Acknowledgments}
This research was conducted at Surlamer Investments. The work, including all findings and results presented in this paper, is the property of Surlamer Investments.

\section{References}

\bibliographystyle{plain}
\bibliography{References}

\begin{thebibliography}{10}

\bibitem{10825449}
Ali Elahi and Fatemeh Taghvaei.
\newblock Combining financial data and news articles for stock price movement prediction using large language models.
\newblock In {\em 2024 IEEE International Conference on Big Data (BigData)}, pages 4875--4883, 2024.

\bibitem{lewis2021retrievalaugmentedgenerationknowledgeintensivenlp}
Patrick Lewis, Ethan Perez, Aleksandra Piktus, Fabio Petroni, Vladimir Karpukhin, Naman Goyal, Heinrich Küttler, Mike Lewis, Wen tau Yih, Tim Rocktäschel, Sebastian Riedel, and Douwe Kiela.
\newblock Retrieval-augmented generation for knowledge-intensive nlp tasks, 2021.

\bibitem{lin-2004-rouge}
Chin-Yew Lin.
\newblock {ROUGE}: A package for automatic evaluation of summaries.
\newblock In {\em Text Summarization Branches Out}, pages 74--81, Barcelona, Spain, July 2004. Association for Computational Linguistics.

\bibitem{mahfouz2021framework}
Mahmoud Mahfouz, Armineh Nourbakhsh, and Sameena Shah.
\newblock A framework for institutional risk identification using knowledge graphs and automated news profiling.
\newblock {\em arXiv preprint arXiv:2109.09103}, 2021.

\bibitem{msci_gics}
{MSCI}.
\newblock The global industry classification standard (gics\textregistered).
\newblock \url{https://www.msci.com/indexes/index-resources/gics}.
\newblock Accessed: 2025-08-31.

\bibitem{pei2024modeling}
Jiaxin Pei, Soumya Vadlamannati, Liang-Kang Huang, Daniel Preo{\c{t}}iuc-Pietro, and Xinyu Hua.
\newblock Modeling and detecting company risks from news.
\newblock In {\em Proceedings of the 2024 Conference of the North American Chapter of the Association for Computational Linguistics: Human Language Technologies (Volume 6: Industry Track)}, pages 63--72, 2024.

\bibitem{ranaldi2024eliciting}
Leonardo Ranaldi, Marco Valentino, and Andr{\'e} Freitas.
\newblock Eliciting critical reasoning in retrieval-augmented language models via contrastive explanations.
\newblock {\em arXiv preprint arXiv:2410.22874}, 2024.

\bibitem{wang2025rag}
Yu~Wang, Shiwan Zhao, Zhihu Wang, Ming Fan, Yubo Zhang, Xicheng Zhang, Zhengfan Wang, Heyuan Huang, and Ting Liu.
\newblock Rag+: Enhancing retrieval-augmented generation with application-aware reasoning.
\newblock {\em arXiv preprint arXiv:2506.11555}, 2025.

\bibitem{yu2024rankrag}
Yue Yu, Wei Ping, Zihan Liu, Boxin Wang, Jiaxuan You, Chao Zhang, Mohammad Shoeybi, and Bryan Catanzaro.
\newblock Rankrag: Unifying context ranking with retrieval-augmented generation in llms.
\newblock {\em Advances in Neural Information Processing Systems}, 37:121156--121184, 2024.

\bibitem{zhang2024raft}
Tianjun Zhang, Shishir~G Patil, Naman Jain, Sheng Shen, Matei Zaharia, Ion Stoica, and Joseph~E Gonzalez.
\newblock Raft: Adapting language model to domain specific rag.
\newblock {\em arXiv preprint arXiv:2403.10131}, 2024.

\bibitem{zhang2020bertscoreevaluatingtextgeneration}
Tianyi Zhang, Varsha Kishore, Felix Wu, Kilian~Q. Weinberger, and Yoav Artzi.
\newblock Bertscore: Evaluating text generation with bert, 2020.

\bibitem{zhao2025alphaagentslargelanguagemodel}
Tianjiao Zhao, Jingrao Lyu, Stokes Jones, Harrison Garber, Stefano Pasquali, and Dhagash Mehta.
\newblock Alphaagents: Large language model based multi-agents for equity portfolio constructions, 2025.

\bibitem{zhou2024finrobotaiagentequity}
Tianyu Zhou, Pinqiao Wang, Yilin Wu, and Hongyang Yang.
\newblock Finrobot: Ai agent for equity research and valuation with large language models, 2024.

\end{thebibliography}
\newpage
\appendix

\section{Queries and Prompts}

\begin{tcolorbox}[colback=gray!5, colframe=black!50, title=Risk Query, fonttitle=\bfseries]

Major Risks of this Company:

Risks are factors related to a company that the performance of the company can be most reliant on; factors that can determine the performance eg: Strategic Supplier Dependence, Tariff and Trade Policy Sensitivity, Customer \& Revenue Concentration, Geographical Concentration, Geopolitical/Regional Exposure, Energy Transition \& Technology Investment, Supply Chain Fragility, Cybersecurity and Digital Risk, Capital Allocation / Financial Structuring, Regulatory/Legal Complexity, Human Capital \& Succession, Macroeconomics/currency correlation, number of the suppliers, regions of activities, Any specific customer, Any specific products, informative actions and events that can cause change in the stock price or future of the company.

\label{prompt:riskquerry}
\end{tcolorbox}

\begin{tcolorbox}[colback=gray!5, colframe=black!50, title= Risk Information Extraction Prompt, fonttitle=\bfseries]

You are an expert-level equity analyst with deep expertise in 
\pinkbold{industry}. 
I am a hedge fund portfolio manager retrieving information for an investment committee meeting. 
You will be given a section of a 10-K/10-Q file, Earning Call Transcripts, or Analyst Reports, 
and you should retrieve related information about a given query.\vspace{2mm}

Company Info : \pinkbold{name}, Ticker: \pinkbold{ticker}, Industry: \pinkbold{industry}\vspace{2mm}

Data: \pinkbold{cosine similarity extracted text chunk}\vspace{2mm}

Task: List the Major Risks of this Company.

Description of risks: Risks are factors related to a company that the performance of the company can be most reliant on; factors that can determine the performance eg: Strategic Supplier Dependence, Tariff and Trade Policy Sensitivity, Customer \& Revenue Concentration, Geographical Concentration, Geopolitical/Regional Exposure, Energy Transition \& Technology Investment, Supply Chain Fragility, Cybersecurity and Digital Risk, Capital Allocation / Financial Structuring, Regulatory/Legal Complexity, Human Capital \& Succession, Macroeconomics/currency correlation, number of the suppliers, regions of activities, Any specific customer, Any specific products, informative actions and events that can cause change in the stock price or future of the company.\vspace{2mm}

Return a list of key phrases as specific intrinsic risks (not general and market risks) and explain why they can trigger and cause a risk. Or why is it informative?\vspace{2mm}

Do not generate any information that is not included in the given text. Do not use prior knowledge; only extract / retrieve and structure relevant information. Avoid any additional descriptions. State the most relevant knowledge from the text based on the given question. Avoid generic and general answers, and too broad answers. Be specific about the company and the industry. Report any information that can be useful from an investment perspective within the given query scope.

\end{tcolorbox}

\begin{tcolorbox}[colback=gray!5, colframe=black!50, title= Aggregation Prompt, fonttitle=\bfseries]

You are an expert-level equity analyst with deep expertise in \pinkbold{industry}. I am a hedge fund portfolio manager retrieving information for an investment committee meeting. Your inputs include summaries of SEC filings, analysts' reports, and earnings call transcripts. You need to answer a question or provide information about the given query based on the given summaries and retrieved information. You should select and aggregate relevant and trustworthy answers and construct a well-rounded analysis on the given query. The length of the answer should depend on what was asked.\vspace{2mm}

Company Info : {{name}}, Ticker: \pinkbold{ticker}, Industry: \pinkbold{industry}\vspace{2mm}

Question: \pinkbold{question}\vspace{2mm}
                              	 
Analysts answers: \pinkbold{data}\vspace{2mm}

The given data is retrieved from different sources of information about the given query. Try to select information from various sources and do not rely only on each of the sources (SEC filings, analysts' reports, and earnings call transcripts). State the sources as much as possible.
Take into account the industry of the company and see broadly, and do not give generic and general answers. Try to connect the information together to come up with new observations about the question that can be informative for investment purposes.

\end{tcolorbox}

\begin{tcolorbox}[colback=gray!5, colframe=black!50, title= Contrasticve Risk Identification Prompt, fonttitle=\bfseries]

Here is a list of major risks for 
\pinkbold{sub-sector} companies. Your task is to generate a risk summary for \pinkbold{target\_company\_name} (\pinkbold{target\_company\_ticker}), using comparative insights from the other companies in the same industry:

You are receiving information for all companies simultaneously, so you should identify risks specific to \pinkbold{target\_ticker} in contrast to the others. Avoid generic or universally applicable risks; instead, highlight how such risks manifest uniquely for \pinkbold{target\_ticker}.

Here is the risk information:\vspace{2mm}

\pinkbold{target\_company\_name} (\pinkbold{target\_company\_ticker}): 

\pinkbold{Risks for target company}\vspace{2mm}

\pinkbold{Peer Company 1} (\pinkbold{peer\_company\_ticker}): 

\pinkbold{Risks for peer company}\vspace{2mm}

\pinkbold{Other Peers Information} \vspace{2mm}

Only give the risks for the company 
\pinkbold{target\_company\_name} (\pinkbold{target\_company\_ticker}). Focus on company-specific, non-generalized/generic insights. Choose the most 3--5 important risks that drive the company's performance.
Your tone should be technical but smooth, including valid reasons and arguments. Also include sources of information and numerical backup only if available and necessary.
\end{tcolorbox}


%
\end{document}